\documentclass[submission,copyright,creativecommons,hidelinks]{eptcs}
\usepackage[utf8]{inputenc}
\providecommand{\event}{IJCAI-ECAI AREA Workshop 2022}
\usepackage{underscore}
\usepackage{hyperref}
\usepackage{lipsum}
\usepackage{csquotes}
\usepackage{graphicx}
\usepackage{multirow}
\usepackage{makecell}

\title{The Need for a Meta-Architecture for Robot Autonomy}
\author{
Stalin Muñoz Gutiérrez \qquad
Gerald Steinbauer-Wagner
\email{
sgutierr@ist.tugraz.at \qquad
steinbauer@ist.tugraz.at
}
\institute{
Autonomous Intelligent Systems Group.\\
Institute of Software Technology.
Graz University of Technology.
Austria.}
}
\def\titlerunning{The Need for a Meta-Architecture for Robot Autonomy}
\def\authorrunning{S. Muñoz Gutiérrez, G. Steinbauer-Wagner}

\begin{document}
\maketitle

\begin{abstract}
Long-term autonomy of robotic systems implicitly requires dependable platforms that are able to naturally handle hardware and software faults, problems in behaviors, or lack of knowledge. 
Model-based dependable platforms additionally require the application of rigorous methodologies during the system development, including the use of correct-by-construction techniques to implement robot behaviors. 
As the level of autonomy in robots increases, so do the cost of offering guarantees about the dependability of the system. 
Certifiable dependability of autonomous robots, we argue, can benefit from formal models of the integration of several cognitive functions, knowledge processing, reasoning, and meta-reasoning. 
Here, we put forward the case for a generative model of cognitive architectures for autonomous robotic agents that subscribes to the principles of model-based engineering and certifiable dependability, autonomic computing, and knowledge-enabled robotics.
 
\end{abstract}

\section{Introduction}
\label{sec:introduction}
It can be argued that \emph{acting} is the crux of robotics research.
It is the agency of robots manifested as continuous interaction with a physical world that distinguishes the robotics field from other scientific and technological disciplines. 
Because robots are designed by humans with the help of computers to fulfill a purpose, they extend our agency.
The quest for robot autonomy is bound to the quest for ours. 
We ponder the implications of intelligent robots able to make not only decisions but reason about their own goals and intentions.
In particular, for safety-critical applications we would like to have guarantees about the behavior of autonomous robots.
Research on the dependability of engineering solutions has a long tradition that has produced a plethora of methodologies, techniques, technologies, standards, and regulations.
We expect from our robots the same we expect from other machines, devices, and tools:
to satisfy the highest quality standards at a minimum cost.
If we are to see autonomous robots potentiate our agency in the world, we should aim to make them certifiable.
However, to many researchers, the mere mentioning of words such as standards and certification brings the preconception that these aspects do not qualify as research activities.
Whether this might be the case when seen from the perspective of specialized research areas, we argue that this judgment does not apply for integrative multidisciplinary endeavors such as \emph{long-term robot autonomy} (LTRA).
Community efforts to come to an agreement on standard models and platforms pay in the long run better than scattered unconcerted attempts to tackle a problem as big as that of LTRA.

A wide diversity of research interests is currently driving the robotics community in an expanding effort to include more aspects of intelligence to enable autonomy in robots.
Nevertheless, integrating autonomous behavior out of a diverse collection of paradigms, formal methods, models, technologies, and methodologies is a heuristic, often trial and error process that does not scale well to complex domains. 
Reference architectures are helpful, provided they can be effortlessly adapted by roboticists to achieve different levels of autonomy. 
In practice, however, implementing a very abstract architecture requires of many design decisions and significant human programming effort prone to interpretation and implementation defects.
More importantly, research within the field of robotics should not be primarily a software engineering effort that do not create novel theories nor ask fundamental scientific questions.

We agree with the view that \emph{Model-based engineering} (MBE) can reliably support the development of highly complex systems that require of safety guarantees. 
The maturity and availability of formal methods, modeling languages, and automation toolchains within the MBE field are no longer limiting factors for a paradigm shift from hand-crafted robotic systems to correct-by-construction affordable certifiable solutions.
In the long run, the research community would likely benefit from interoperability enabled by MBE models. 
Models that would not only be executable facilitating their use by practitioners, but also amenable to be easily reproduced, analyzed, compared, and improved by any research group that wishes to do so. 

The contribution of our work is an integrative literature review that spans different concerns relevant to LTRA. 
We put forward the case for a generative model of cognitive architectures, that is, a formal description captured in one or more \emph{domain specific modeling languages} (DSMLs) enabling automatic synthesis of concrete architectures for LTRA as a satisfying solution to a set of traceable requirements.
We will highlight the need to further study the integration of several deliberation functions besides acting, that are necessary for intelligent behavior. 
In addition, we consider that cognitive penetrability and meta-reasoning models should be explicit and formal, amenable to theoretical analysis, verification, and automatic transformations. 
In the next section, we comment on relevant research on dependability, model-driven engineering, autonomic computing, autonomy, and robot architectures. Section \ref{sec:architecture} will present the case for a meta-architecture for autonomy in robots. 
Finally, in Section \ref{sec:conclusion}, we present our concluding remarks and future work.

\section{Related Work}
\label{sec:related-work}
We start our presentation by highlighting the relevant qualities of dependable systems because we consider dependability to be concomitant to LTRA.  
In our account, we intentionally left out many core aspects that apply to almost all engineering application domains such as fault tolerance, reconfiguration, graceful degradation, and so on. Instead, in the interest of conciseness, we will focus our presentation on deliberative and high-level cognitive functions.

\subsection{Dependability}

Traditional studies on the conceptualization of dependable information systems include the works of Laprie~\cite{laprie:1985} and Avi\^{z}ienis et al.~\cite{avizienis:laprie:randell:landwehr:2004}. 
Their concepts and taxonomies cover a wide range of engineering concerns to take into consideration when defining a generic cognitive architecture that aims for dependability. 
For Avi\^{z}ienis et al., dependability is a \emph{global concept} encompassing several attributes: availability, reliability, safety, integrity and maintainability. Closely related concepts seeking equivalent goals are dependability or the ``ability to deliver service that can justifiably be trusted [and]... to avoid [unacceptable] system failures",  
\emph{high-confidence} where ``consequences of the system behavior are well understood and predictable'',
\emph{survivability} or the ``capacity of a system to fulfill its mission in a timely manner'', and \emph{trustworthiness} or the ``assurance that a system will behave as expected''.
According to their study, dependability can be engineered by means of four mechanisms: (1) \emph{fault prevention}, (2) \emph{fault tolerance}, (3) \emph{fault removal}, and (4) \emph{fault forecasting}. They highlight the importance of \emph{self-checking}, implying that the mechanisms responsible for handling faults, need to be themselves protected from the impact of the potential faults.

The European SPARC 2020 Robotics Multi-Annual Roadmap~\cite{SPARC:2020} (MAR) proposes the following levels of dependability numbered from 0 to 7 in increasing order: \emph{no dependability}, \emph{mean failure dependability}, \emph{fails safe}, \emph{failure recovery}, \emph{graceful degradation}, \emph{task dependability}, \emph{mission dependability}, and \emph{predictive dependability}. 
A generative cognitive architecture would need to address all levels above level 2 as far as the robot behavior is concerned. 
The roadmap also identifies \emph{cognitive dependability} as significant for robots that require understanding, interpretation, and reasoned decision making. 

Dependability of robotic systems needs to be  \emph{certified}. 
A certificate resolves that the deployment of an autonomous system cannot incur in undesirable consequences nor lead to unacceptable risks based on pondering \emph{claims} and \emph{evidence}, and \emph{argumenting} in favor of the veracity of the claims, given the evidence~\cite{rushby:2008}.
The \emph{Committee on Certifiably Dependable Sofware Systems} (CCDSS) of the National Research Council of the National Academies~\cite{national:2007} defines certification as ``the process of assuring that a product or process has certain stated properties, which are then recorded in a certificate.'' 
To certify a software as dependable they consider it should be done based on a ``credible'' \emph{dependability case}~\cite{weinstock:goodenough:hudack:2004} (a generalization of the concept of \emph{safety case}~\cite{bishop:bloomfield:1998}). 
The required level of dependability may significantly constrain the architecture, models and technologies chosen for its implementation. 
Some additional key points made by the committee are that developing software for dependable systems is difficult and costly and that reliance on testing does not constitute sufficient evidence if high dependability is required (evidence from analysis is also necessary). 
About the last point, extensive testing is nevertheless mandatory. 
Unit tests, fuzz testing (randomly generated tests), automatic test generation from models such as state machines (model-based testing), invariants and run-time assertions , as well as performing automatic tests after every maintenance action constitute effective techniques towards achieving dependability requirements. 
Some of the existing standards are: 
 (1) \emph{Common Criteria} (CC), a security certification standard for software; 
 (2) \emph{Software Considerations in
Airborne Systems and Equipment Certification} (RTCA DO-178B)(in Europe is known as ED-12B),  a worldwide standard for software in civilian aircraft, and 
(3) International Electrotechnical Commission's (IEC's) 61508 (in US its counterpart is known as ISA S84.01), a standard for functional safety of electrical/electronic/programmable electronic-safety-related systems.

Addressing all relevant aspects of dependable systems is beyond the scope of this presentation, it requires processes that span beyond the strictly technical realm.
For example, CCDSS points out the need to evaluate a dependability claim beyond the technical dimension. 
It is also necessary to evaluate the organization that stated the claim, to assess the integrity of the evidence chain. 
As for the importance of a \emph{safety culture}, the committee regards it as important to grant organizational support to personnel attitudes aiming for the highest quality standards.
In~\cite{steinbauer:loigge:muhlbacher:2016}, Steinbauer et al. propose a \emph{Model-based Development Process} (MBDP) for dependable autonomous systems. 
This model covers the whole life-cycle of the system as well as different aspects of the system architecture.
A comprehensive treatment of this approach including how to deal with principal threats to dependable systems can be found in~\cite{steinbauer:2016}.

\subsection{Model Based Engineering}

Engineering solutions benefit extensively from the use of models to synthesize solutions to real-world problems faced by society.
Within the field of software engineering, modeling has been there since the beginning. 
Recently, the term
\emph{Model-Driven Engineering} (MDE) has emerged as a promising approach that can effectively cope with the increasing complexity of software systems. In~\cite{schmidt:2006}, Schmidt presents a clear perspective on why MDE is, at least in intent and focus, different in the role that models play within the field. 
The main distinction is that most abstractions and traditional models have primarily been \emph{computing-oriented} as opposed to \emph{domain-oriented}.
He interprets the limited adoption of \emph{Computer-Aided Software Engineering} (CASE) methods by the software engineering community as an example that illustrates that models by themselves are not sufficient. 
According to his assessment, one of the reasons for the arguable failure of CASE is that the models on which CASE relied were too abstract, with a \emph{``one-size-fits-all''} perspective that created enormous challenges for automatic translation to particular system platforms, therefore creating big \emph{semantic gaps} between the design intent and its often intricate realization using specific technologies. To address this problem, Schmidt proposes developing MDE technologies by combining two approaches: (1) 
\emph{Domain Specific Modeling Languages} (DSMLs), and (2) transformation engines and generators. 
DSMLs are meta-models tailored for the application domain. 
The purpose of DSMLs is to allow developers to express design intent \emph{declaratively} and not \emph{imperatively}. 
The purpose of transformation engines and generators is to ensure consistency with functional and non-functional requirements.
Additionally, MDE has to deal with the orchestration and customization of product-line architectures, model checkers, aspect-oriented languages, design patterns, etc.

In~\cite{kent:2002}, Kent discusses the benefits of a \emph{model-centric} approach for architecting engineering solutions.
A primary reason for having an architectural description is to be able to communicate effectively with different human stakeholders. 
Models abstract away platform-specific concerns. 
From the perspective of the system development, they allow for manipulation and transformation by machines facilitating automatic validation of correctness and model-driven testing.
He reflects on the potential of \emph{meta-models}, envisioning their use to drive a set of automated tools where tools can generate other tools and reconfigure themselves. 

Steck et al.~\cite{steck:lotz:schlegel:2011} applied MDE to the development lifecycle of service robot architectures. 
In their work, they define robotic meta-models for software components. 
Models then ease their transformation into executable code and are used for various analyses and simulations at design-time, and support decision-making at run-time. 
Models are a means to describe the features of components, allowing the application of Component-Based Software Engineering (CBSE) principles~\cite{brugali:scandurra:2009,brugali:shakhimardanov:2010}. 
The central element of their architecture is a \emph{model pool} that contains various models with heterogeneous representations required for the particular technological platforms and component implementations. 
Meta information makes it possible to query information, retrieve them in the target representation or make it possible to reason about them.
Their toolchain is \textsc{SmartSoft}~\cite{smartsoft:2009}, a robotic framework that instantiates their platform independent meta-model \textsc{SmartMARS}.

In~\cite{stampfer:lotz:lutz:schlegel:2016}, Stampfer et al. present SmartMDSD toolchain, an integrated model-driven software development environment (DE) for robotics software development based on \textsc{SmartSoft}. 
One of the main features of SmartMDSD is that it integrates several DSMLs into a single consistent DE. 
Its design is based on finding the right trade-off between \emph{freedom of choice} and \emph{freedom from choice}. 
The former aims not to enforce any decisions while the latter provides sufficient guidance regarding the composability and system-level conformance. 
Additional embraced design principles are \emph{separation of roles}, \emph{separation of concerns}, and \emph{composability and composition}.
RobMoSys~\cite{espinoza:jahn:principato:2020} is a MDE approach to robotics development. 
It relies in meta-models incorporating Papyrus~\cite{radermacher:morelli:hussein:nouacer:2021} for Robotics and SmartMDSD. RobMoSys tackles complexity by considering separation of concerns: (1) computation, (2) communication, (3) coordination, and (4) configuration.
Concerns apply across different levels of abstraction: (1) hardware, (2) OS System/Middleware, (3) execution container, (4) function, (5) service, (6) skill, (7) task plot, and (8) mission. 
The project aims for robot technologies to be composable, modular, predictable, safe, reusable, and certified. An integrated feature of Papyrus is the possibility of automatically generating hazard and risk analysis.
There are currently several robotic platforms following MDE principles, a recent survey can be found in~\cite{de-araujo-silva:valentin:hughes-carvalho:da-silva-barreto:2021}.

\subsection{Autonomic Computing}
One example of the increasing importance of self-managing computer systems is the IBM \emph{autonomic computing initiative}~\cite{ganek:corbi:2003}. 
The term autonomic was chosen based on the autonomic nervous system in animals that releases the conscious brain from duties that are vital (heart rate, respiratory rate, digestion, etc.) but monotonous.
The initiative supports the case for the development of systems that are: 
(1) 
\emph{self-configuring}, systems automatically adapt to dynamically changing environments, new features and capabilities can be added/subtracted/enabled/disabled;  
(2)
\emph{self-healing}, systems detect, diagnose, and react to anomalies, they predict problems and take preventive actions, their design maximizes reliability and availability; 
(3)
\emph{self-optimizing}, the system can monitor and adapt resources automatically optimizing across multiple heterogeneous systems; and 
(4)
\emph{self protecting}, they anticipate, detect, identify, and protect themselves from attacks. 
The report also calls for the widespread use of open industry standards.

As pointed out by Garlang et al.~\cite{garlan:cheng:huang:schmerl:2004} the need to cope with changing environments points towards systems that are capable of dynamic \emph{self-adaption} by means of external mechanisms. 
He remarks that although self-adaptation is being in computer systems for a while, the mechanisms are often specific to the application, the code has no separation of concerns and exhibits only localized handling of faults.
External self-adaptation mechanisms subscribe to proper software engineering principles where separation of concerns and loose coupling allow for modules than facilitate analysis, maintenance, modification, and reuse by new or existing modules or systems. 
There is a case for \emph{architectural models} (AMs) to enable reasoning about self-adaptation for satisfying system-level behaviors and properties, as well as integrity constraints. 
They proposed the \emph{Rainbow} framework, an abstract AM and reusable infrastructure to support self-adaptation. 
The framework monitors the properties of a running system, checking for constraint violations against the AM that consists of a graph made of \emph{components} (computational elements) linked by \emph{connectors} (their interactions). 
Rainbow models the system adaptation by means of an extended \emph{architectural style} that includes the notion of \emph{adaptation operators} (AOs) and \emph{adaptation strategies} (ASs).
AOs (for example adding or removing a service) are possible actions to perform by the control infrastructure to change the configuration of the system. 
ASs are adaptations necessary to avoid an undesirable condition (for example to activate a graceful degradation in the services).

Oreizy et al.~\cite{oreizy:gorlick:taylor:hembigner:johnson:medvidovic:quilici:rosenblum:wolf:1999} defines software as \emph{self-adaptive} if it modifies its own behavior in response to changes in the operating environment. 
In deciding the self-adaptation capabilities of a system, several concerns must be taken into consideration: (1) triggering conditions; (2) open or close adaptation, i.e. it is open if the system can incorporate new behaviors while running, it is close otherwise; (3) type or level of autonomy; (4) cost-effectiveness of adaptation; (5) frequency of adaptation; and (6) information required for deciding on triggering adaptation. 
\index{self-adaptive systems!self-adaptability spectrum}
According to the \emph{spectrum} of self-adaptability proposed by them, highly adaptive solutions will exhibit a clear separation of software-adaptation concerns from other software functions. 
\index{dependability!in self-adaptive systems}
Safety, reliability, and correctness are primordial for self-adaptive systems design. 
It is also important to guarantee the consistency and integrity of the system. 
Oreizy et al. propose the use of an \emph{Architecture Evolution Manager} (AEM) that mediates all adaptations. 
The AEM enforces that all changes are \emph{atomic}, meaning they should not fail, and if they do so they should not be deployed, leaving the application unaltered. Another fundamental architectural component in their proposal is the Adaptation Management (AM). 
AM is a monitor of the application and its environment. 
It enables adaptations, plans them, and deploys change descriptions to the running system. 
These functions can be implemented using independent software agents. 
For decision making, observations are gathered from exceptional events generated by \emph{inline observers} that check relevant assertions. 
Monitoring of \emph{patterns} of events may be necessary to trigger an adaptation, this can be implemented using expectation agents. 
Consistency Management (CM) can check \emph{invariants} offline on \emph{attributed graph grammars} that capture all possible configurations for the application, in this representation graph rewrites represent architectural changes. 
Observers performing runtime monitoring on the application and the environment are employed to detect inconsistencies against relevant \emph{annotations}.

\subsection{Autonomy}

Franklin and Graesser~\cite{franklin:graesser:1997} discuss the attributes of autonomous agents and provide a general definition that applies to solutions and systems within the field of robotics: 
\enquote{An \emph{autonomous} agent is a system situated within and a part of an environment that senses that environment and acts on it, over time, in pursuit of its own agenda and so as to effect what it senses in the future.}
As pointed out by the proponents, this embracing definition equally comprises complex sophisticated cognitive agents as it does a simple closed-loop controlled device such as a thermostat. They identify additional capabilities that are more suitable for an operational definition in the context of decision making, the core features are reactivity, autonomy, goal orientation, and temporal continuity. 
In the context of robotics research, Tessier~\cite{tessier:2017} quotes the definition of autonomy considered by the Defense Science Board as \enquote{the capability to independently compose and select among different courses of action to accomplish goals based on its knowledge and understanding of the world, itself, and the situation}~\cite{defense-science-board:2016} (Originally from Shattuck~\cite{shattuck:2015}). 
Although highly relevant to the discussion, and in particular to clarify why a concept of \emph{dependable autonomy} is not a \textit{contradictio in terminis}, we are not addressing here Decisional Autonomy from the perspective of \emph{shared authority} and the broader view that autonomy can be understood as a power relationship between human and robot agencies. 
Module this abstraction, autonomy is, in words of Castelfranchi and Falcone, \enquote{due to agent's architecture}~\cite{castelfranchi:falcone:2003}. The argument for a cognitive architecture is therefore of necessity not of sufficiency.

\emph{Decisional autonomy} is an important topic covered in the MAR.
The level of autonomy of a robot is a consequence of its cognitive abilities but it is also parameterized by: \emph{environmental factors} modulating its empowerment, \emph{decision cost} demanding risk-aware well informed choices, the \emph{time scale} of deployments demanding high dependability, and the extent of the \emph{decision range} making the required level of autonomy more complex is it widens.
Hereafter when discussing autonomy we omit the adjective \emph{long-term}, as this can be understood as a cognitive ability parameter that demands the robot a higher level of dependability.

\subsection{Robot Architectures}

We acknowledge that robot architectures are diverse and accept numerous variations.
Our intention here is not to give a comprehensive account nor to put forward a new taxonomy. 
We bring into attention some fundamental concepts and a handful of reference architectures chosen because of their prominence and their influence in the research community.
We intend to bring to the table, through these exemplars, the concepts that have proven essential and operational to real-world robotic solutions. 
It should not be implied that other works are less influential or important.

\subsubsection{Software Engineering}

Software engineering theory and good practices of architectural models are structural to the equation. 
The realization of any paradigm, theory, behavior, or experimental study in robotics requires a software engineering dimension.
Architectural descriptions \emph{are models} described in architectural description languages (ADLs).
In practice, architectural descriptions go all the range from just a collection of connected boxes without underlying semantics, to fully formal models.

In~\cite{medvidovic:taylor:2000}, Medvidovic and Taylor identify core principles of software architectures and provide a conceptual framework that we briefly summarize here.
Architectures focus on the high-level structure of a software application.
ADLs allow the explicit modeling of components, connectors, configurations, constraints, hierarchical composition, computational paradigms, and communication paradigms among others.
The first concern of ADLs is that of expressivity and \emph{representation}.
They also must assist the whole design process, facilitate static and dynamic analysis, and address evolution, refinement, traceability, and simulation.
Toolsets come together with an ADL, their adoption depends strongly on the availability of a flexible and powerful set of technologies that facilitate the realization of the architecture during the whole life-cycle of development.
One important aspect to consider when selecting an ADL beyond its concrete syntax is the semantic theory that subsumes the semantic framework that comes with it. 
Some of these theories are: statecharts, finite state machines,
partially-ordered event sets, chemical abstract machines, process algebras, Petri nets, algebraic or axiomatic formalisms.
The adherence to semantic theories facilitates the automatic tools to perform model transformations, analysis, evaluation, and verification for example.
Three fundamental elements of software architectures are (1) components, (2) connectors, and (3) architectural configurations.
For the first two elements, the ADL will allow its formal definition, and specification of interfaces, types, semantics, constraints, evolution, and the so-called non-functional properties among which more dependability aspects are expressed (safety, and reliability for example).
Regarding architectural configurations, it includes the definition of a \emph{topology}, a graph of components and connectors representing the architectural structure in a way that allows for the assessment of concurrent and distributed aspects. 
The qualities of the configuration description are understandability, compositionality, refinement, traceability, and heterogeneity.
Qualities of the described system are heterogeneity, scalability, evolvability, and dynamism.
Just to highlight one aspect, dynamism considers the modifications to the architecture where the system is running, this is important for safety- and mission-critical robotic systems.

ROS~\cite{quigley:conley:gerkey:faust:foote:leibs:wheeler:ng:2009} has become the default choice for implementing software architectures in robotic solutions.
In a recent study, Malavolta et. al.~\cite{malavolta:lewis:bradley:lago:garlan:2020} analyse the adoption of ROS by the robotics community. 
In their study, they report on the use of ROS by practitioners, the top architecturally supported capabilities found are control, planning, vision, and navigation. 
The top quality requirements are maintainability, reliability, performance, and usability. 
A set of 49 architectural guidelines based on evidence were identified by their study.

\subsubsection{Autonomy}

Robot architectures specialized for autonomy have also been successfully implemented. In~\cite{alami:raja:fleury:ghallab:ingrand:1998}, Alami et al. present an integrated architecture consisting of three levels: \emph{decision}, \emph{execution}, and \emph{functional}.  
They consider that autonomous robots must exhibit deliberation and reactive capabilities, broadly encompassing:
(1) decision-making, 
(2) reactivity, 
(3) situation anticipation
and (4) context-awareness.
In their architecture, the decision level is both goal- and event-driven and includes automated planning, plan execution supervision, and responsiveness to events generated in other layers.
The execution level, also known as \emph{executive} is responsible for controlling and coordinating the execution of functions while satisfying the requirements of the task at hand.
The functional level corresponds with the built-in robot action and perception capabilities.
For them, deliberation is an important faculty, they define it as \enquote{both a goal-oriented process wherein the robot anticipates its actions and the evolution of the world, and also, a time-bounded, context-dependent decision-making process for a timely response to events.}
Their implementation of the architecture includes the use of the temporal planner IxTeT, the Procedural system for task refinement and supervision PRS, the reactive control of the functional level Kheops, and GenoM for the specification and integration of modules at the functional level.
PRS is a sophisticated implementation of the Belief-Desired-Intention paradigm as well as an expressive formal language that is a trade-off between declarative and imperative programming~\cite{ingrand:2014}. 
GenoM is a Model-Driven Software Engineering Tool that automatically generates the module's code as much as possible.
A description language allows the definition of a given module capturing: (1) the services that it manages, (2) the input and output parameters, (3) associated execution code fragments and their real-time characteristics (time period and delay, for example), and (4) an exhaustive list of possible execution reports (feedback on success or failure of a module execution).
Alami et al. architecture for autonomy has remained a reference architecture among the robotics research community. 
In a recent work~\cite{hladik:ingrand:dal-zilio:tekin:2021}, Hladik et al. present a Model-based execution engine for the control and verification of real-time systems. 
This engine developed at LAAS-CNRS is part of a \emph{toolchain} called \emph{HIPPO} that supports a unified modeling language for design, verification, and execution.
\index{model-driven engineering!programming languages!G\textsuperscript{en}\textsubscript{o}M}
HIPPO allows for the automatic generation of both executable and verifiable code from a high-level specification based on the robotic programming framework G\textsuperscript{en}\textsubscript{o}M. 
The generated code is interpreted by a dedicated \emph{real-time} execution engine.

Another archetypical example is the \emph{Remote Agent} (RA) architecture for autonomous spacecrafts of NASA of Muscettola et al.~\cite{muscettola:nayak:pell:williams:1998}. 
Remote autonomous agents require of the following characteristics: (1) computational intelligence that exhibit high-levels of autonomy and continuous operation over long periods of time, (2) guarantees of meeting tight deadlines and resource constraints, (3) high reliability under partial observability and in the event of faults, and (4) precise coordination of concurrent processes over several subsystems.
The three principles implemented in RA architecture are: (1)
the use of \emph{compositional declarative models} and \emph{model-based programming}, (2) \emph{on-board deduction and search}, and (3) \emph{high-level closed-loop control}. 
Rather than relying on simplifying assumptions and abstractions, or depending on human supervision and decision-making, RA uses declarative models for the synthesis of appropriate responses to anomalies and unexpected scenarios. RA performs searches in a highly \emph{tuned}, propositional, best-first kernel.
Their high-level control design allows the declarative definition of missions, instead of the traditional concrete low-level commands defined for nonautonomous operations.
There are four main architectural components of RA: (1) The \emph{Reactive Executive} (EXEC), (2) The temporal \emph{Planner/Scheduler} (PS), (3) the \emph{Mission Manager} (MM), and (4) the \emph{Model-based mode Identification and Reconfiguration} (MIR) module. 

Here we highlight only what we consider main dependability dimensions addressed by MIR.
In~\cite{williams:nayak:1996}, Williams and Nayak further elaborates on the MIR module, presenting \emph{Livingstone} the kernel for an autonomous \emph{model-based} diagnostic system.
Livingstone uses fast propositional conflict based algorithms for model-based diagnosis combined with a propositional, conflict-based feedback controller. 
Their work relies in a central \emph{single model} for a variety of autonomous tasks including detecting anomalies, performing diagnosis, fault recovery, and fault avoidance. 
Components of the system are modeled using concurrent transition systems specified using temporal logic. Models are qualitative by choice, aiming to minimize detailed modeling mistakes that may lead to erroneous diagnoses. To achieve reactivity, their system limits reasoning to consider only the present and next state of the system.
In~\cite{nayak:bernard:dorais:gamble-jr:kanefsky:kurien:millar:muscettola:rajan:rouquette:smith:taylor:tung:1999}, Nayak et al. provides with a detailed report on the formal verification of RA. 
The verification included nominal operation, goal-oriented commands, closed-looped plan execution, failure diagnosis, recovery, replanning, as well as system-level fault protection.
Because the testing of RA autonomous operation would require of an impractically large number of test cases, and also due to the limited availability of high-fidelity test beds that were unable to run faster than real time, a \emph{baseline testing} approach was conducted combined with the use of \emph{lower fidelity testbeds} resenbling a \emph{pyramid}.

\subsubsection{Cognition}

In~\cite{reiter:2001}, Reiter states that 
\enquote{Cognitive Robotics has, as its long-term objectives, the provision of a uniform theoretical and implementation framework for autonomous robotic or software agents that reason, act and perceive in changing, incompletely known, unpredictable environments.}

In a recent survey~\cite{kotseruba:tsotsos:2020}, Kotseruba and Tsotsos review over 40 years of research in cognitive architectures identifying 84 of them with over different 900 projects. 
They provide a classification based on knowledge representation (KR) and information processing that divides them into \emph{emergent} (E) and \emph{hybrid} (H).
Emergent architectures are based on massively parallel models, within this group, a further classification divides them into \emph{connectionist logic systems} (CLS) and \emph{neural modeling} (NM).
Hybrid architectures combine symbolic and sub-symbolic approaches.
A further refinement of this category divides them into \emph{fully integrated} (FI), \emph{symbolic sub-processing} (SSP), and purely \emph{symbolic} (S).
The most cited works include ACT-R (in H,FI), Soar (in H,FI), CLARION (in H,FI), ICARUS (in H,S), EPIC (in H,S), and LIDA (in H,FI). 
The previously presented architecture PRS is also classified as symbolic (in H,S).
Discussing the particulars of these cognitive architectures is out of the scope of this integrative review.

Higher cognitive abilities enable higher levels of autonomy.
MAR identifies the following cognitive abilities for robotics applications: (1) action, mainly concerned with planning and acting; (2) interpretative, or the ability to make sense of the environment and other agents; (3) envisioning, or the ability to understand the consequences of its actions and relevant events occurring in the environment; (4) learning, concerned with knowledge acquisition and (5) reasoning, or the capacity to process knowledge to make rational conclusions and choices. 

In~\cite{wang:baciu:yao:kinsner:chan:zhang:hameroff:zhong:hunag:goertzel:2010}, Wang et al. present perspectives on \emph{cognitive informatics} and \emph{cognitive computing} (CC).
In particular, they report on CC as an emergent paradigm aiming to develop \emph{cognitive computers}\cite{pylyshyn:1988} with the capacity to think and learn applicable to autonomous agent systems. 
They consider cognitive computers as \enquote{as knowledge processors beyond those of data processors in conventional computing}. 
These computers would enable the simulation of \emph{machinable thought} that consists on several reasoning modes: inferences, problem-solving, decision making, etc. 
They identify as a major challenge towards the realization of cognitive computers the handling of inconsistent information. 
The concept of \emph{cognitive penetrability}\cite{pylyshyn:1988} relates closely to the plasticity of the human cognitive process: \enquote{what we know determines how we react}. 
The problem of modeling how to overcome inconsistencies is interesting not only because of the necessary adequate handling of them, but because according to cognitive science they may trigger important learning processes. 
For example, Zhang and Grégoire~\cite{zhang:gregoire:2016} describe that \enquote{through the process of overcoming an inconsistency, the agent is able to revise, refine, or augment its existing knowledge to adapt to the emergent patterns and behaviors as exhibited in the inconsistent circumstance}. Their work also reports on several effective techniques for handling inconsistency in intelligent systems.

Beetz et al.~\cite{beetz:mosenlechner:tenorth:2010} develop the Cognitive Robot Abstract Machine (CRAM), a software platform for the development of \emph{cognition-enabled} autonomous robots. 
They identified the need for software tools for the implementation of high-level cognitive abilities such as learning and knowledge processing. 
The CRAM kernel includes CRAM plan language(CPL) specially designed for mobile manipulation and cognition-enabled control. 
The design principle is that control programs are not reduced to code artifacts but also exist as entities that allow reasoning on them. 
Another element of the kernel is \emph{KnowRob} a knowledge processing system based on the formalism of description logic~\cite{baader:2003,van-Harmelen:lifschitz:porter:2008} (see Section~\ref{lab:sec-knowledge-enabled}).
On top of their kernel, there is \emph{COGITO}, a meta-reasoning component that reasons and deals with plan execution flaws.
\emph{Cognitive extensions} (CRAM-EM) allow for: (1) action awareness, (2) learning and adaptation, (3) transformational planning, and (4) web-enabled robot control.

In~\cite{ingrand:ghallab:2017}, Ingrand and Ghallab survey deliberation for autonomous robots. In their review, six deliberation functions are relevant for autonomous robots: (1) planning, (2) acting, (3) monitoring, (4) observing, (5) goal-reasoning, and (6) learning. 
The importance of deliberation resides in its expression as deliberative acts or acting towards intended objectives.
They acknowledge the importance of bounded rationality and the need for meta-reasoning functions that trade deliberation time for action time according to the current goals and constraints.
The integration of planning and acting is critical. 
Models used for planning are descriptive (\emph{know-what}) while models necessary for acting are operational (\emph{know-how}).
Ingrand and Ghallab classify robot architectures in four groups: (1) hierarchical, (2) reactive, (3) teleo-reactive, and (4) open. 
Hierarchical architectures are the most common choice. 
Components are organized into layers having different levels of abstraction. 
They identify as a challenge the implementation of deliberation functions that span multiple levels.
Reactive architectures have limited or no need for deliberation.
They are built based on input-output mappings implemented by automata.
Teleo-reactive architecture integrates planning and acting along different levels of abstraction using a unified representation. 
A primary concern of these approaches is their scalability.
Open architectures are suitable for complex open environments that require of knowledge processing capabilities. 
These architectures allow a robot to integrate models and knowledge from the web.

\begin{figure}
    \centering
    \includegraphics[width=0.53\textwidth]{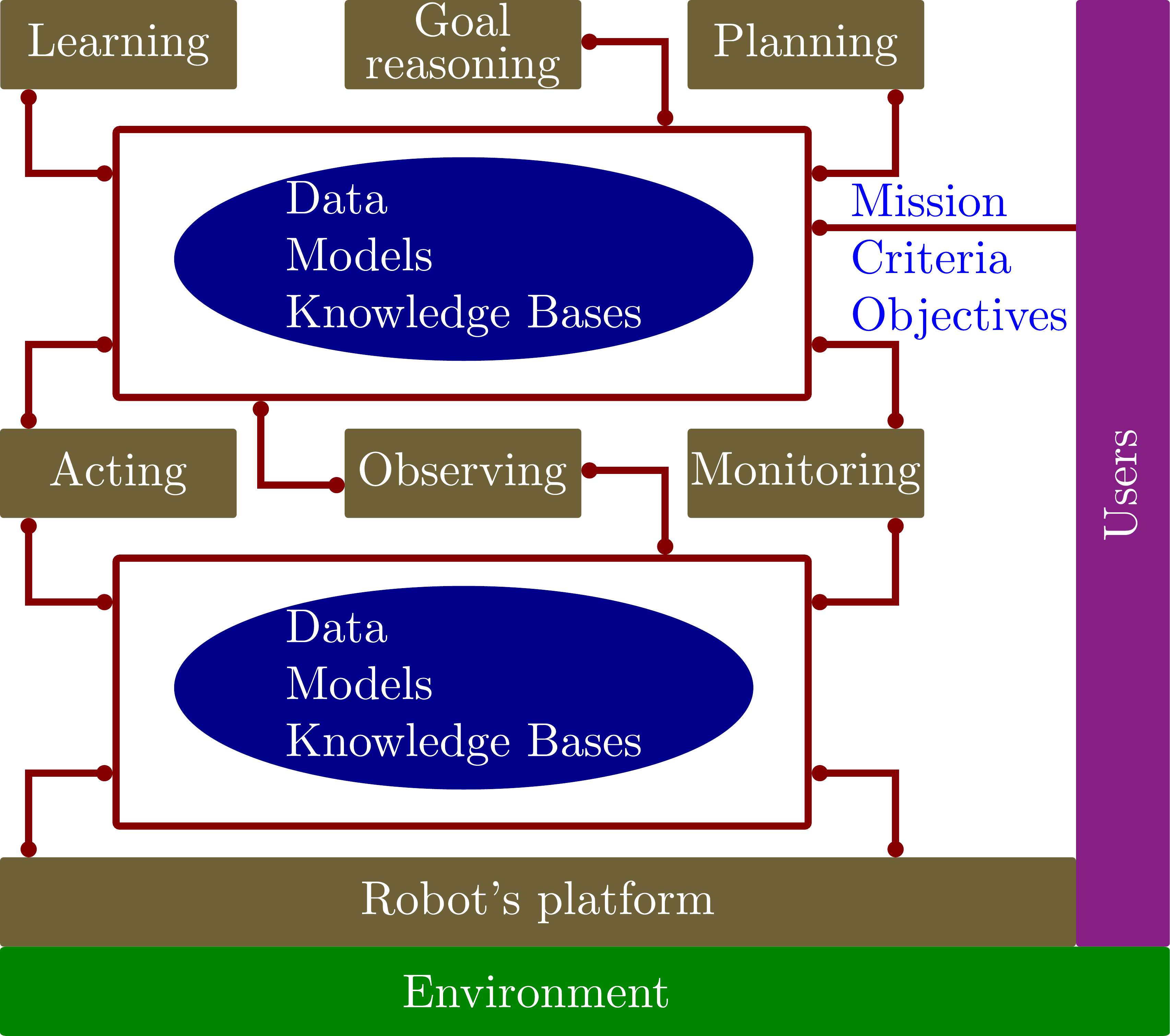}
    \caption{Depiction of a typical hierarchical architecture implementing several deliberation functions (modified from~\cite{ingrand:ghallab:2017}). At the level of deliberation, the robot platform encapsulates and abstracts away the complexity of sensory-motor functions of the robot.
    More than one knowledge base might be required across different levels of abstraction.}
    \label{fig:refarchitecture}
\end{figure}

In Figure~\ref{fig:refarchitecture}, a depiction of a hierarchical deliberative architecture based on Ingrand and Ghallab's proposal shows how the different deliberation functions are part of the decisional autonomy of the robot. 
A particular instantiation of this architecture implementing some of the deliberation functions applied to a industrial logistics scenario can be found in~\cite{ulz:ludwiger:steinbauer:2019,de-bortoli:munoz-gutierrez:steinbauer-wagner:2021}. 
In this architecture, there is more than one knowledge base enabling decision-making and reasoning tasks at the different levels of the abstraction hierarchy.
Connectors are represented by buses rather than individual pairs to show the intent of integrative approaches.

In~\cite{cox:mohammad:kondrakunta:gogineni:Dannenhauer:larue:2021}, Cox et al. present MIDCA cognitive architecture. 
MIDCA is based on \emph{computational meta-cognition} processes analogous to an \emph{action-perception cycle}. 
The difference is that inputs are not interpretations of external stimuli but cognition states and instead of acting in the world, actions regulate cognitive activity.
Cox et al. identify three fundamental forms: (1) explanatory, (2) immediate, and (3) anticipatory. Explanatory responds to \emph{failures} in already executed cognitive processes, immediate is a form of runtime introspection, and anticipatory is self-directed foresight.
The purpose of explanatory meta-cognition is to diagnose what caused a cognitive-level failure by formulating a meta-level goal to correct the situation.
Anticipatory meta-cognition is a form of goal reasoning over \emph{suspended goals}, those that the agent could not achieve because of insurmountable constraints.
Meta-cognitive functions appear in many robotic solutions, for example, RA implemented a form of meta-planning.

\subsubsection{Knowledge-Enabled}
\label{lab:sec-knowledge-enabled}
In~\cite{reiter:2001}, Reiter subscribes to the knowledge representation hypothesis (KRH) of Smith~\cite{smith:1982}:
\enquote{
Any mechanically embodied intelligent process will be comprised of structural ingredients that (a) we as external observers naturally take to represent a
propositional account of the knowledge that the overall process exhibits, and (b) independent of such external semantical attribution, play a formal but causal and essential role in engendering the behavior that manifests that knowledge.
}
This hypothesis has dominated the KRR research~\cite{levesque:brachman:1985} for many years. 
To a certain degree (see MAR higher cognitive abilities in the previous section) knowledge is a horizontal concern across all cognitive abilities for robots, especially when high levels of cognitive ability are required. 
The remark here is that knowledge-enabled robots use explicit models, as compared to implicit knowledge embedded within code and data artifacts.

In~\cite{tenorth:beetz:2017}, Tenorth and Beetz study KR for autonomous robots. 
They discuss \emph{KnowRob} a  \emph{Knowledge Processing System} (KPS), advocating for the use of this paradigm to address the problem of filling the gap between information-rich low-level sub-symbolic representations and abstract high-level descriptions. 
To address this problem, their approach engineers common representational structures and appropriate inference procedures that enable posing and resolving queries necessary for the execution of abstract instructions by \emph{interpreting} background knowledge and the context of the task at hand.
An advantage of their approach is that it builds on top of already existing data and algorithmic robotic components by enriching them with semantic annotations. 
They acknowledge the advantages of logical approaches, regardless of whether they take the form of action theories, planning-oriented descriptions, or probabilistic representations, for modeling actions, preconditions, and effects that entail control, prediction, and analysis tasks. 
Unfortunately, most of the time they rise applicability concerns related to high or unpredictable computational costs. 
For real-world applications, the number of atoms in the axiomatizations grows rapidly forcing the modeler to keep a high level of abstraction but requiring the agent to resolve the fine level of detail required for the execution of actions in the environment. 
By preserving the already efficient sub-symbolic and ad hoc data structures and behaviors, there is no impact on efficiency nor information loss due to the abstraction required for symbolic inference.
They refer to this constructs as computing \emph{symbolic views} as needed or \emph{on-demand abstractions}. Rather than attempting to aim for a complete and consistent world model the knowledge representation is a semantic \emph{integration layer}.
Some of the insights from their work are: (1) No fixed levels of abstraction, depending on the use case at hand the right number of abstraction levels differ;
(2) a knowledge base should reuse data structures of the robot's control program; (3) symbolic knowledge bases are useful, but not sufficient, a \emph{shallow} symbolic representation in the knowledge base interpret rich detailed information relevant for appropriate behavior; (4) robots need multiple inference methods, temporal reasoning of events can be combined with ontological reasoning and spatial reasoning, for example; (5) evaluating a knowledge base is difficult, for example, run-time of queries, or scalability concerns, a major challenge resides on the quantification of these aspects.
Key to the knowledge processing system is the use of the W3C web ontology language~\cite{motik:patel-schneider:parsia:bock:achille:haase:hoekstra:horrocks:sattler:2009} (OWL) to represent a \emph{common ontology} that is used as a \emph{glue} to unify several reasoning algorithms. 

Based on web services, other research initiatives have successfully developed platforms that allow robots to query, process, or share knowledge according to the task at hand~\cite{waibel:beetz:civera:d-andrea:elfring:galvez-lopez:haussermann:janssen:montiel:perzylo:schiessle:tenorth:zweigle:de-molengraft:2011,beetz:tenorth:winkler:2015}. Open architectures exploit \emph{ontology-based} approaches to augment the level of autonomy in robots. For a review on the field we refer the reader to~\cite{olivares-alarcos:bessler:khamis:goncalves:habib:bermejo-alonso:barreto:mohammed:rosell:quintas:2019}, where Olivares-Alarcos et al. classify and compare different ontology-based approaches.

\section{Towards a meta-architecture for autonomy}
\label{sec:architecture}

In this section, we connect the evidence presented in previous sections to build the case for a formal cognitive meta-architectural description. 
A meta-architecture is not an architecture but a description of a family of architectures. 
It should not be confused with meta cognition~\cite{cox:2005}, although we consider meta cognition as an important architectural driver.
A meta-architecture is closely related to the concept of architectural templates.

We summarize the analysis dimensions presented in the previous sections. 
LTRA in robots requires of high levels of \emph{dependability}. 
The research within this dimension is highly mature, currently providing effective techniques and industrial standards.
\emph{MDE} can support the life cycle of robotic solutions by means of methodologies and automation of model transformations and code artifacts synthesis. 
Correct-by-construction can significantly reduce the cost of certification. 
Additionally, models facilitate reproducibility and widespread usage of research outputs by practitioners. 
Almost independently of the level of autonomy, \emph{autonomic computing} provides robust and adaptive behavior. 
Explicit architectural modeling of autonomic functions is fully amenable to MDE and vice versa.
Regarding the \emph{software engineering} dimension, we consider that architectures for autonomy should aim to be expressed in ADLs with clear and sound semantic theories that facilitate analysis and runtime verification at the same time aim to reduce the semantic gap for the robotics field.
Specific applications demanding a particular level of \emph{autonomy} will benefit from a meta-architecture that is able to derive the best compromise between \emph{cognitive abilities} and cost.
The robust integration of the different cognitive functions remains the main challenge, and it is a significant threat towards dependable robotic solutions. 
Finally, besides enabling complex behavior and rational decision-making, knowledge-enable robotics contribute towards the trustworthiness of a robotic system operating in complex environments by facilitating its transparency and explainability. 

Regarding cognition, our vision includes, although is not limited to, the possibility of an open community standard model of cognitive computer that can ease the transition to a model-fueled ecosystem. 
There is already important progress in different dimensions necessary to achieve such conceptual and technological platform.
This means that such developments can be built on top of existing paradigms with arguably minimal effort.
One advantage of having an open standard at a meta-model level would mean that there is no strict requirement to collapse the diversity of approaches into a common implementation, instead each particular platform would be free to implement the standard using its own conceptual frameworks and toolchains.

As far as \emph{acting} is concerned, we assess that there are already widely and well-proven execution engines including those provided in RobMoSys, and PRS. 
In particular, RobMoSys is an excellent candidate as a basis for implementing the required cognitive views. 
It is a technologically sound platform from the point of view of software engineering.
One important remark is that the aim here would be to close the semantic gaps that limit the implementation of higher-level cognitive abilities.
CRAM is also a visionary work that is already in the same path.
The formal model execution engine \textsc{Hippo} is addressing the semantic gaps between high-level models and code artifacts through automated code generation based on the semantic theory of Petri nets.
One possible obstacle toward higher cognitive functions allowing complex decision-making and reasoning is that the research effort has so far concentrated on expressive languages directed to humans.
A roboticist has to a great extent solve the \emph{control logic} that then is encoded in an expressive language, whether OP procedures (OpenPRS), behavior trees (RobMoSys), or the formal specification language \textsc{Fiacre} (\textsc{Hippo}).
We should aim to transit, progressively, to action theories.
The control logic should be fully produced by the artificial agent, not a human programmer.

The challenges of integrating multiple cognitive deliberation functions as identified by Ingrad and Ghallab~\cite{ingrand:ghallab:2017} give ample opportunities for research.
It is still an art to come out with effective ways to handle the integration of cognitive functions.
The current abundance of cognitive architectures provides a rich landscape for drawing paradigms of cognition to hitherto mainly engineering-motivated endeavors.
In the field of cognitive architectures, there is already a call for a standard model out of community consensus~\cite{laird2017standard}. 
We consider that a standard model for robot autonomy with higher-level deliberation functions is feasible.
If realized as a formal model using ADLs, it can be grounded by automatic means fully or to a great extent to the currently sound and efficient technologies.
One possible approach is to consider the integration of cognitive functions in the context of meta-reasoning layers.
An approach based on MIDCA architecture would allow modeling cognitive process as mental \emph{acts}. 
The advantage of this unifying approach is that we already have much of the necessary machinery in place. 
There is no apparent barrier using action theories to model cognitive processes. 

Standardized knowledge representation models are not required as long as the appropriate transformation can be perform automatically. 
Meta-models, and consistent conceptualizations are more important. 
Meta-models should be abstract enough to be independent of the robot platform.
As in many cognitive architectures, the skill level can be the base line.
Underlying levels can also be considered, but we grant more importance to the exploitation of existing software modules that are widely used by the robotics community. 
In this context, it is important to adequate integrate dependable solutions out of unreliable components.
Models that include computational complexity descriptions can be exploited by meta-reasoning processes to optimize internal cognitive state and modulate the cognitive load of the robot.

\section{Concluding Remarks}
\label{sec:conclusion}
Meta-models have proven to be effective for enabling high-level cognitive abilities in autonomous robots.
Typically, however their scope do not include the integration of high-level deliberation functions corresponding to an explicit formal cognitive model.
The existing MDE methods and technologies are currently mature enough to support the transition towards a full model-enabled development and operation of robotic solutions (programming-free above the skill level).
Although, we consider important to work towards a standard formal model agreed by the community.
Such standard model can then be automatically transformed into concrete robotic solutions with the possibility of automatically verifying and certifying their dependability.
Our future work will aim at modeling a meta-architecture for LTRA.
\section{Acknowledgments} \label{acknowledgments}
Funded by LEAD project  ``Dependable  Internet  of  Things  in  Adverse Environments'', TU Graz. 

\section{Bibliography}

\bibliographystyle{eptcs}
\bibliography{area}

\end{document}